\pdfoutput=1
\documentclass{article}

\usepackage[preprint, nonatbib]{neurips_2024}

\usepackage[utf8]{inputenc} 
\usepackage[T1]{fontenc}    
\usepackage[pagebackref=true,colorlinks,citecolor=blue,linkcolor=blue,urlcolor=blue,hypertexnames=false]{hyperref}
\usepackage{url}            
\usepackage{booktabs}       
\usepackage{amsfonts}       
\usepackage{nicefrac}       
\usepackage{microtype}      
\usepackage{xcolor}         

\usepackage[most]{tcolorbox}

\usepackage{colortbl}
\usepackage{paralist}
\usepackage{graphicx}

\definecolor{mygray}{rgb}{0.89, 0.93, 0.85}
\definecolor{whitesmoke}{rgb}{0.96, 0.96, 0.96}
\definecolor{timberwolf}{rgb}{0.86, 0.84, 0.82}

\newlength\savewidth
\newcolumntype{x}[1]{>{\centering\arraybackslash}p{#1pt}}

\newcommand{\app}{\raise.17ex\hbox{$\scriptstyle\sim$}}

\newcommand{\ourmethod}{{LLAVADI}}

\newcommand \footnoteONLYtext[1]
{
	\let \mybackup \thefootnote
	\let \thefootnote \relax
	\footnotetext{#1}
	\let \thefootnote \mybackup
	\let \mybackup \imareallyundefinedcommand
}

\title{{\ourmethod}: What Matters For Multimodal Large Language Models Distillation}

\author{Shilin Xu$^{1,3}$ $^{*}$, Xiangtai Li$^{2}$, Haobo Yuan$^{2}$, Lu Qi$^{3}$,  \\ \textbf{Yunhai Tong$^{1}$, Ming-Hsuan Yang$^{3}$} \\
  {$^{1}$PKU}
  {$^{2}$NTU} 
  {$^{3}$UC Merced} \\
  \textit{Email: xiangtai94@gmail.com and xushilin@stu.pku.edu.cn}
}

\begin{document}

\maketitle

\begin{abstract}
%
The recent surge in Multimodal Large Language Models (MLLMs) has showcased their remarkable potential for achieving generalized intelligence by integrating visual understanding into Large Language Models.
Nevertheless, the sheer model size of MLLMs leads to substantial memory and computational demands that hinder their widespread deployment. 
In this work, we do not propose a new efficient model structure or train small-scale MLLMs from scratch. 
Instead, we focus on what matters for training small-scale MLLMs through knowledge distillation, which is the first step from the multimodal distillation perspective. 
Our extensive studies involve training strategies, model choices, and distillation algorithms in the knowledge distillation process. 
These results show that joint alignment for both tokens and logit alignment plays critical roles in teacher-student frameworks. In addition, we draw a series of intriguing observations from this study. 
By evaluating different benchmarks and proper strategy, even a 2.7B small-scale model can perform on par with larger models with 7B or 13B parameters. Our code and models will be publicly available for further research.

\end{abstract}

\section{Introduction}
\label{sec:intro}

Multi-modal large language models (MLLMs) show promising potential for building general assistants due to their visual perception, reasoning, and localization ability. Most research works focus on expanding the instruction data, model architecture, or training strategy to fulfill the cross-modality properties of MLLMs.
%
The state-of-the-art LLaVA~\cite{liu2023llava} model designs a data engine to unify numerous vision-language tasks, including visual question answering, conversation, complex reasoning, and captions. 
It uses a projection layer to convert visual tokens into the language model's feature space. These visual tokens, along with language tokens, are then inputted into LLMs for joint processing by the language models. The training process is supervised using the next-token prediction paradigm.
Many methods have been developed to enhance various aspects of LLaVA, including improving the alignment between visual and textual representations, optimizing the projection layer efficiency, and tackling challenges in integrating multimodal data. These advancements aim to enhance the model's performance. Some approaches have also expanded LLaVA into video domains or added grounding and segmentation capabilities, enabling a broader range of applications for multimodal tasks.

On the other hand, enabling cross-modality capacities in resource-constrained scenarios is also required for application purposes. Most existing MLLMs are 7B or 13B models, and most applications cannot support such large models. 
Therefore, research efforts are increasingly focused on developing more efficient model architectures and compression techniques that retain cross-modality capabilities while reducing computational and memory requirements. These approaches include model pruning and quantization, which aim to make MLLMs more accessible and practical for deployment. Some methods aim to design efficient modules. For example, MobileVLM~\cite{chu2023mobilevlm} proposes an efficient projector connecting visual and language tokens. FastV~\cite{chen2024image} and VTW~\cite{lin2024boosting} suggest reducing or eliminating visual tokens to decrease computational complexity. However, these studies do not leverage the knowledge of existing stronger MLLMs, which we show can be transferred to smaller MLLMs.

In this work, we explore efficient MLLMs from a distillation perspective. Existing LLM-based distillation works mainly aim for language task distillation, with no visual inputs as condition tokens. Thus, directly applying these works may not lead to improvements or even inferior results (See Sec.~\ref{sec:exp}). On the other hand, existing multimodal distillation mainly focuses on specific settings, including image retrieval and visual question answering on specific datasets. To our knowledge, there are no works to explore the extensive studies on MLLMs despite the previous studies that have designed various distillation methods. As the first step in this direction, we raise one essential question, "\textit{What are the most effective aspects of MLLM distillation?}"

To answer this question, we conduct an extensive empirical study from four aspects to study the knowledge distillation between stronger teacher models and student models, including feature embedding distillation, logit-level distillation,  affinity-aware distillation, and data-driven knowledge distillation. 
Our empirical study is motivated by the previous distillation methods in LLMs, multi-modal learning, and image representation learning domains.
Feature embedding distillation explores the intermediate hidden states between the teacher and student models. Logit-level distillation is a basic distillation format on the LLM's classification logits.  Affinity-aware distillation explores the affinities of visual and language tokens, which push students to learn the visual-language relationship of the teacher model. Data-driven knowledge distillation explores the distribution gap between training data and the initial student model.

Through our studies, we observe several interesting results. (1) Feature level distillation can also improve the student results and work orthogonal to logit level distillation. However, it is more sensitive to layer position in LLMs. (2) Naive logit level distillation works well, even better than specifically designed LLM distillation loss.  (3) Affinity-aware distillation losses do not work well in various cases and we think affinity distillation is incompatible with the autoregressive loss. (4) Data-driven knowledge distillation can improve student results by adding teacher-generated and instruct-tuning data. However, the training cost is large due to the data regeneration process. Combining these studies, we term our framework LLAVA DIstillation ({\ourmethod}), a simple yet efficient distillation framework for MLLM distillation, by joint distilling features and logits, which are refined by teacher-generated data and instruction tuning data. 
We conduct extensive studies on various teacher and student baselines and show that {\ourmethod} can generalize to numerous teacher and student frameworks.
%
The main contributions of this work are: 
\begin{itemize}
\item We explore the first detailed studies on MLLM distillation from four different aspects, where our goal is to answer what matters in MLLM distillation.
\item We show that most existing LLM distillation methods cannot bring extra gains for MLLM, while simple logit and feature distillation achieve sufficient improvements. 
\item We demonstrate that adding teacher-generated data and instruction-tuning data can improve performance.
\item We demonstrate the superiority and efficiency of {\ourmethod} in multimodal datasets on various baselines. 
\end{itemize}
\section{Related Work}
\label{sec:related_work}
%

\noindent
\textbf{Knowledge Distillation.} Staring from image classification tasks, the majority of research~\cite{hinton2015distilling,guan2020differentiable,yuan2020revisiting,zhang2020distilling,xie2020self} in knowledge distillation focuses on designing better distillation loss functions. 
The first work~\cite{hinton2015distilling} transfers knowledge from large models to small ones. 
Subsequently, numerous works~\cite{xie2018improving,liu2019structured,shu2021channel,wang2020intra,zhang2020improve,wang2019distilling} explore the knowledge distillation on various tasks, including vision tasks and language tasks, where they adopt task-specific designs.  
Recently, several works~\cite{agarwal2023onpolicy,wen2023fdivergence,zhu2024llavaphi,alpaca,vicuna2023, minillm} mainly study large language model distillation methods. 
%
%
The most common knowledge distillation method for large language models involves minimizing output distribution discrepancies between the teacher and student models using \textit{forward} KL divergence.
However, if a student model is unable to learn all aspects of a highly complex teacher, the resulting behavior known as "mode-covering" may cause the student to assign probability to tokens with low probability in the teacher's distribution. This mode-covering effect can result in hallucinations and produce low-quality outputs.
Several methods~\cite{minillm, agarwal2023onpolicy, wen2023fdivergence} utilize reverse KL divergence. This prioritizes tokens where the teacher assigns high probabilities to prevent students from overestimating low-probability areas in the teacher's distribution. However, these methods may also face a "mode-seeking" issue, often resulting in reduced diversity.

Other approaches~\cite{sanh2019distilbert,tinybert,liang2022taskaware} align the intermediate hidden states of the student model with those of the teacher. 
For close source large language models like GPT series~\cite{brown2020language, Achiam2023GPT4TR}, some studies~\cite{wang2022selfinstruct,ding2023enhancing,mftcoder2023} aim to transfer knowledge from these models by augmenting the training data.
In contrast, {\ourmethod} emphasizes a multimodal large language model that produces outputs closely aligned with visual inputs.

\noindent
\textbf{Multimodal Model Distillation.} Few early works~\cite{fang2021compressing,wang2022multimodal} have studied the cross-modal distillation on specific tasks, including VQA and caption. 
DistillVLM~\cite{fang2021compressing} performs transformer distillation using hidden attention distribution and feature maps with MSE loss. 
MAD~\cite{wang2022multimodal} aligns visual and text token features between teacher and student with selected tokens. 
Recently, several works~\cite{tinyclip, yang2023clipkd} have studied the distillation of CLIP models. 
In particular, TinyCLIP~\cite{tinyclip} adopts affinity mimicking and weight inheritance to improve small CLIP models, while CLIP-KD~\cite{yang2023clipkd} finds simple feature distillation can achieve good results.
%

\noindent
\textbf{Multimodal Instruction-Tuning.} Early works~\cite{li2023blip2,Alayrac2022FlamingoAV,awadalla2023openflamingo,liu2023llava,zhu2023minigpt} have proposed various multimodal model architectures. 
The standard designs have a unified transformer architecture, vision encoder, and language models. 
With the rise of LLMs, many works focus on combining visual knowledge into LLMs. The fusion of multimodal information can improve various downstream tasks, such as image captioning, visual question answering, and multimodal summarization.

LLaVA~\cite{liu2023llava} is a seminal work that directly feeds visual tokens into LLMs, allowing them to grasp nuanced image information and perform intricate image-based reasoning.

%
%
After LLaVA, lots of works~\cite{chen2023sharegpt4v, ye2023mplugowl, Qwen-VL,dai2023instructblip, cai2024internlm2} apply visual instruction tuning under various applications.
Recently, several works~\cite{chu2023mobilevlm,Yuan2023TinyGPTVEM} have explored efficient MLLMs to fit the urgent demand for deployment.
MobileVLM~\cite{chu2023mobilevlm} explores 1B/3B vision-language models under resource-constrained scenarios.
LLaVA-Phi~\cite{zhu2024llavaphi} achieves stronger results in visual reasoning tasks via Phi–2.7B models.
Recently, several works~\cite{qi2024generalizable, hanoona2023GLaMM} have also explored dense predictions with MLLMs.
%
%
In contrast to prior arts, {\ourmethod} explores knowledge distillation on multimodal large language models and tries to find what components matter. 
To the best of our knowledge, we are the first to carry out such extensive studies.
\section{Method}
\label{sec:method}
%
    

\subsection{{\ourmethod} Framework}

\noindent
\textbf{Overview of LLaVA.} 
The LLaVA~\cite{liu2023llava} architecture consists of a vision encoder, a projection layer, and a large language model. The vision encoder $g$ is initialized from the pre-trained CLIP visual encoder ViT-L/14~\cite{CLIP}, which outputs the visual feature $Z_v \in \mathbb{R}^{HW/P^2\times D_v}$ for a given image $X_v \in \mathbb{R}^{H\times W\times 3}$. The variables $H$ and $W$ represent the height and weight of the image, while $P$ and $D_v$ indicate the patch size and the hidden size of the visual feature. The projection layer adopts a trainable projection matrix $W$ that converts the visual feature into the space of language embedding:
\begin{equation}
H_v=W\cdot Z_v, \text{with}\medspace Z_v=g(X_v).
\end{equation}
The LLM $f_\phi(\cdot)$ parameterized by $\phi$ adopts the architecture of LLaMA~\cite{touvron2023llama}, which accepts the instruction text token $H_q$ and aligned visual feature $H_v$ as input. The final response $Y_A$ is supervised using an autoregressive manner as follows,
\begin{equation}
\label{equ:autogressive}
p(Y_a|H_v, H_q) = \prod_{i=1}^{L}p(y_i|H_v, H_{q,y<i}).
\end{equation}
The LLaVA model uses a two-stage training consisting of pre-training and instruction fine-tuning stages. In the first pre-training stage, only the projection layer is trainable, and it learns to align the visual feature into the space of language embedding. This stage involves 558K image-text pairs filtered from CC3M, and each image-text sample can be treated as a single-turn conversation. The instruction fine-tuning stage keeps the visual encoder frozen and maintains the projection layer and LLM trainable.

\begin{figure}
    \centering
    \includegraphics[bb=0 0 1676 960, width=\textwidth]{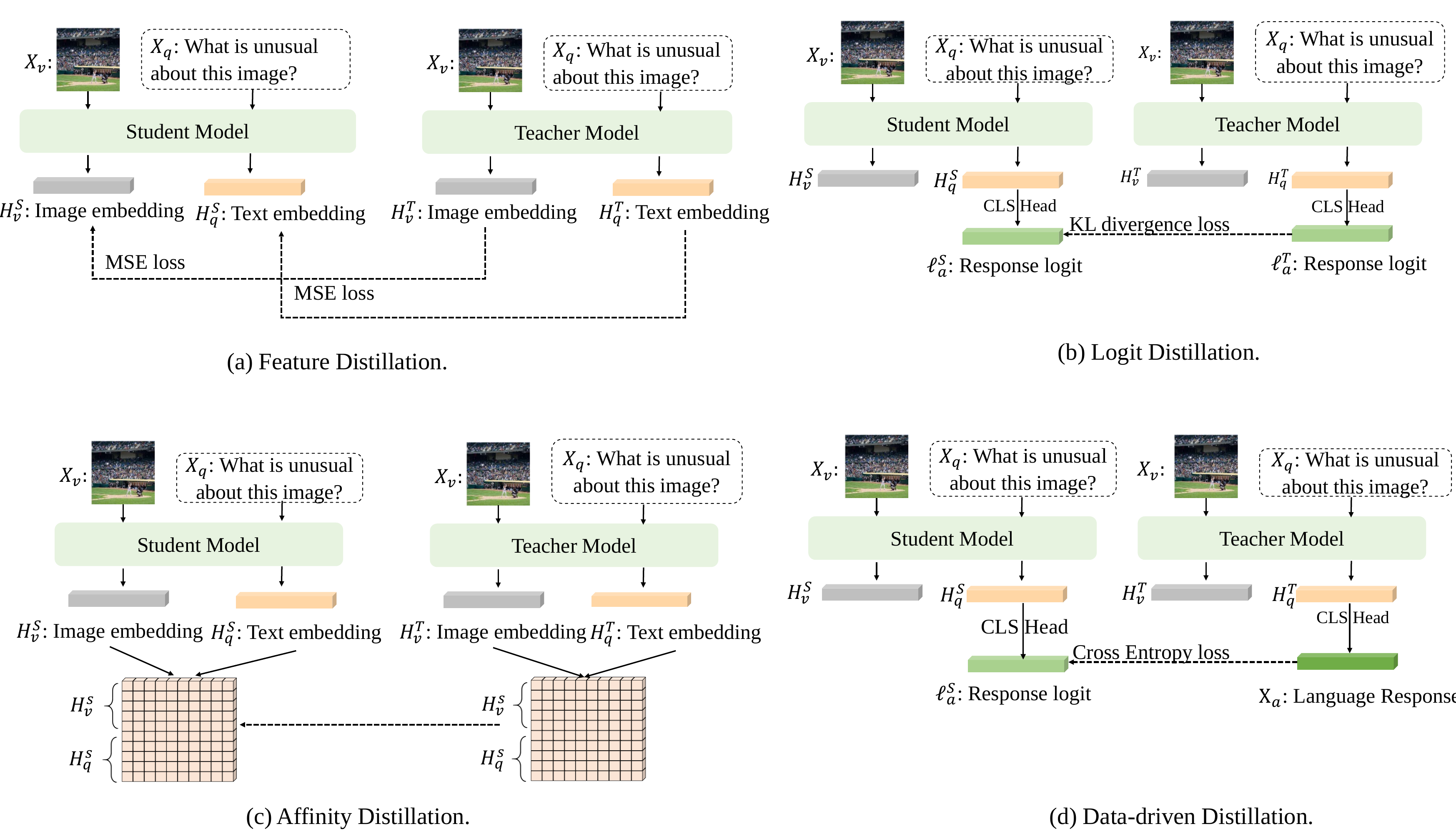}
    \caption{Illustration of various knowledge distillation approaches proposed in this work.}
    \label{fig:teaser}
\end{figure}

\noindent
\textbf{{\ourmethod} Architecture.} 
In Fig.~\ref{fig:teaser}, we illustrate the various knowledge distillation approaches proposed in this work. It contains feature distillation, logit distillation, affinity distillation, and data-driven distillation. All these approaches consists of a teacher model $T$ and a student model $S$.
While both models follow a similar MLLM architecture, the student model has significantly fewer parameters than the teacher model. Specifically, the student model features fewer attention layers and smaller feature dimensions or a reduced visual encoder than the teacher model. Our choice for the teacher model is LLaVA-v1.5-13B~\cite{liu2023llava}, while the student model employs MobileLLaMA~\cite{chu2023mobilevlm} as its LLM, which only contains 2.7B parameters.
The visual encoder and projection layer of the student model are consistent with the teacher model, as the proportion of parameters in these two modules is relatively small. Additionally, our student model undergoes two-stage training exclusively, and we solely employ the distillation process in the fine-tuning stage and the supervisor. 

\subsection{Exploration on What Matters in MLLM Distillation}

\noindent
\textbf{Overview.}
We first discuss the methods used for training the student model. We employed various KL divergences for logit distillation and aligned the hidden embeddings between the teacher and student models. Additionally, we experimented with different data augmentation techniques for supervision fine-tuning.

\noindent
\textbf{Knowledge Distillation on Feature Embeddings.}
Layer-wise distillation~\cite{hinton2015distilling,Zuo2022MoEBERTFB,Sun2019PatientKD,tinybert,Sun2020MobileBERTAC,Hou2020DynaBERTDB,liang2022less} is a commonly used technique to transfer knowledge from a teacher model to a student model. As shown in Fig.~\ref{fig:teaser}(a), feature embeddings distillation involves distilling feature embeddings $H$ from the intermediate transformer~\cite{Vaswani2017AttentionIA} layer of the teacher model to corresponding layers of the student model. Here, $H$ denotes the output feature embeddings for each transformer layer, which can be divided into image embedding $H_v$ and text embedding $H_q$. These methods utilize different similarity metrics to measure and enhance the alignment of internal representations between the two models, facilitating a broader transfer of hidden information. The goal of this approach is to ensure that the student model generates similar outputs as the teacher and handles information in a comparable way at hidden layers.

Due to the inconsistency in the feature dimensions between the teacher model and the student model, we employ a simple MLP layer to map the features of the student model to the feature space of the teacher model. We refer to the training objective of these hidden embeddings as \textbf{$L_{cos}$}, which can be implemented using either \textit{cosine embedding} or \textit{MSE} loss.

\noindent
\textbf{Knowledge Distillation on Logit}
Fig.~\ref{fig:teaser}(b) shows the architecture of knowledge distillation on logit. 
Aligning the logit between the teacher and student models directly is a regular distillation approach. Following the previous methods~\cite{hinton2015distilling, sanh2019distilbert}, we adopt a \textit{softmax-temperature} based KL distillation:
\begin{equation}
    L_{ce} = D_{KL}(P_t||P_s) = \sum_{c \in \mathcal{C}} P_t(c)\cdot\frac{P_t(c)}{P_s(c)}, \text{with}\medspace p_i=\frac{\exp \left(z_i / T\right)}{\sum_j \exp \left(z_j / T\right)}
\end{equation}
Here, the $z$ means the logits obtained from the classification head of LMM, $\mathcal{C}$ means the total number of vocabulary size of our LMM, and $T$ represents the temperature, influencing the smoothness of the output distribution. Higher $T$ values increase randomness, while lower values enhance determinism by favoring the most probable words. The same temperature $T$ is used for both student and teacher during training. 
In specific implementation, we refer to $D_{KL}(P_t||P_s)$ as \textbf{forward KL} while $D_{KL}(P_s||P_t)$ as \textbf{reverse KL} due to the asymmetry of KL divergence. To balance the forward KL and reverse KL, we also implement \textbf{generalized JSD} (Jensen-Shannon divergence). The \textbf{JSD}($\beta$) introduces a parameter $\beta$ to interpolate between the forward and reverse KL divergences, resulting in a more flexible divergence measure.
\begin{equation}
    JSD_{KL}(\beta) = \beta D_{KL}(P_t||\beta P_t + (1-\beta) P_s) + (1-\beta)D_{KL}(P_s||\beta P_t + (1-\beta) P_s)
\end{equation}


\noindent
\textbf{Knowledge Distillation on Affinity Leaning.}
Current MLLMs integrate visual features from the vision encoder into LMMs for response generation through next-token prediction. The teacher model is more powerful, able to understand visual features more accurately and capture the relationship between visual and textual features well. Therefore, the teacher model can generate more reasonable and accurate text answers based on image content and questions. If the teacher and student models use the same vision encoder, their image features should be the same. The only difference is the relationship between image features and text features. We refer to the relationship between different features as affinity, and the student model can mimic the affinity distribution of the teacher model to enhance its comprehension of visual features.

\noindent
\textbf{Data Driven Knowledge Distillation.}
Compared to traditional knowledge distillation algorithms, data augmentation has emerged as a prevalent paradigm for achieving knowledge distillation of LLMs. The most advanced LLMs~\cite{ChatGPT,GPT4, team2023gemini} treated as the teacher model can generate a large amount of context-rich and domain-specific training data. The student model trained with these generated data samples can closely mimic the intricate understanding and cognitive abilities of the teacher model. In addition to generating data from teacher models, some methods~\cite{sdft,Agarwal2023OnPolicyDO,ko2024distillm} suggest generating data samples using the student model to bridge the ability gap between it and the teacher model. The logits from the teacher model for these newly generated data samples are used as supervision for $L_{ce}$.

\section{Experiments}
\label{sec:exp}

\noindent
\textbf{Datasets.} Following LLaVA~\cite{liu2023llava}, we mainly use six datasets for comparison, including GQA  \cite{hudson2019gqa}; SQA$^\text{I}$(ScienceQA-IMG~\cite{lu2022learn}), VQA$^\text{T}$(TextVQA~\cite{singh2019towards}), POPE~\cite{li2023evaluating}, MME$^\text{P}$(MME Perception \cite{fu2023mme}), and MMB$^\text{dev}$(MMBench-dev~\cite{liu2023mmbench}). We mainly use VQA$^\text{T}$, SQA$^\text{}$, MME$^\text{P}$, and POPE for empirical studies and report final results on all benchmarks.

\noindent
\textbf{Teacher and Student Models.} If not specified, we will use LLaVA-v1.5-13B~\cite{liu2023llava} as the teacher model for our experiments. The teacher model LLava-V1.5-13B~\cite{liu2023llava} uses the pre-trained CLIP~\cite{CLIP} ViT-L/14 with only 0.3B parameters as its visual encoder. 
The visual encoder is frozen during both the pre-training and fine-tuning stages. 
Our student model has the same visual encoder as the teacher model, and unless stated otherwise, it remains frozen during the distillation process. 
The teacher model utilizes the large language model Vicuna-13B-v1.5~\cite{vicuna2023}, while our student model employs the MobileLLaMA~\cite{chu2023mobilevlm} with 2.7B parameters as our large language model. MobileLLaMA has a similar architecture to Vicuna, but with fewer transformer layers and smaller feature dimensions. 
Thus, the student model shares the same vocabulary size as the teacher model and can seamlessly align the logits between the teacher and student model. 

\noindent
\textbf{Implementation details.} Similar to LLaVA-v1.5~\cite{liu2023llava}, we pre-train the student model on image-text pairs datasets and then fine-tune it with instruct-tuning datasets.
Specifically, we mainly explore the distillation process in the instruct tuning stage since the pre-training stage only involves one project layer. 
Like LLaVA, we freeze the vision encoder and LLM and focus on only training the projector during the pre-training stage. We trained the projector on CC-595K~\cite{liu2023llava} for one epoch using a batch size of 256 and a learning rate of 1e-3. For fine-tuning, we used the 665K instruction-tuning dataset from LLaVA~\cite{liu2023llava} with a batch size of 128 and a learning rate of 2e-5 unless otherwise specified. The projection layer and LLM are learnable during the second fine-tuning stage. 
We use the AdamW~\cite{Loshchilov2017FixingWD} optimizer without weight decay and a cosine learning rate schedule with a warmup ratio of 3\% throughout both stages of training. We only apply the distillation methods for the fine-tuning stage.

Before knowledge distillation, we train our baseline model using CLIP ViT-L/14 visual encoder and MobileLLaMA-2.7B as our LLM. The training strategy of the baseline model follows the LLaVA-v1.5~\cite{liu2023llava}.

\subsection{Feature level distillation}
\label{exp_sec:token_level}
Layer-wise distillation is a powerful tool to compress knowledge from teacher models into smaller ones. This method involves transferring information at each layer of the teacher model to the corresponding layer of the student model. By aligning intermediate representations, layer-wise distillation ensures that the student model not only mimics the final output of the teacher model but also captures the hierarchical features and patterns learned at each stage of the network. However, under general settings, the student model has a smaller transformer layer and fewer feature dimensions than the teacher model. This discrepancy in depth and dimension poses a challenge for direct layer-to-layer alignment. To address this, We only consider aligning the last few layers between the student and teacher models as we think the last few layers contain more complex high-level representation. Then, we employ a learnable MLP layer that projects the hidden embeddings from the student model into the feature space of the teacher model, and the MLP layer will be discarded during inference. We evaluate various feature-level knowledge distillation methods in Tab.~\ref{tab:feature}. The findings indicate that aligning only the last layer yields better results than aligning the last two layers, which leads to significantly worse performance than the baseline.  We think that the reason may be that there is a significant difference in the distribution of intermediate hidden representations between the teacher model and the student model, and simply imitating the hidden layer of the teacher model can lead to performance degradation. The hidden representations of the last layer contain high-level semantic information crucial for token classification.
It is beneficial to align these final layer representations between the teacher and student models. 

Another interesting finding is that aligning all tokens can lead to  slight performance improvement than aligning only answer tokens $Y_a$. This conclusion is opposite to the one in logit-level distillation.
\begin{table*}[h]
    \centering
    \setlength{\tabcolsep}{4mm}{\begin{tabular}{c|cccc}
    \toprule
    Method & VQA$^\text{T}$ & SQA$^\text{I}$ & POPE & MME$^\text{P}$ \\
    \midrule
    \rowcolor{whitesmoke} Baseline & 47.5 & 59.7 & 85.2 & 1307.7 \\
    Last layer & 48.3 & 58.6 & 86.0 & 1268.7 \\
    Last layer + All tokens & 48.2 & 59.3 & 86.4 & 1286.4 \\
    Last two layers & 36.2 & 47.7& 82.5& 1015.8 \\
    \toprule
    \end{tabular}
    }
    \caption{Feature level distillation. We will align intermediate representations of the transformer layer between the teacher and student model. }
    \label{tab:feature}
\end{table*}

\begin{tcolorbox}[
colframe=black,
boxsep=1pt,
]
    \paragraph{\textbf{\text{Finding} 1.}} Only the hidden embeddings of the last layer are effective, while aligning the last two layers achieves much worse results.
\end{tcolorbox}

\subsection{Logit level distillation}
\label{exp_sec:feature_level}

In Tab.~\ref{tab:logit}, we investigate different logit alignment methods for knowledge distillation. The default temperature coefficient of 0.7 is used for all methods during training. The baseline is trained just using autoregressive supervision as shown in Equ.~\ref{equ:autogressive}, and the logit level distillation is an extra supervision during the fine-tuning stage. 
We also use different KL divergence or MSE loss to align the classification logits between the teacher and student models. 
The results indicate that the MSE loss performs significantly worse than the baseline across all evaluation benchmarks, suggesting its unsuitability for logit alignment. On the other hand, KL divergence consistently delivers superior or comparable results in all aspects.  For KL divergence, we try to use methods such as forward KL, reverse KL, and generalized JSD, which are mentioned above in our experiments. The generalized JSD with a parameter $\beta$=0.5 produces slightly improved results for VQA$^\text{T}$ and SQA$^\text{I}$. 
This indicates that using balanced KL divergence is more effective in distilling logits when dealing with a large vocabulary size, enabling the transformation of complex distributions from the teacher and student models. In addition to exploring different KL methods, we also try to use the logit standardization method following~\cite{Sun2024Logit}. 
The logit standardization method results in a significant decrease of MME$^\text{P}$ while improving by 0.6 and 1.2 for VQA$^\text{T}$ and SQA$^\text{}$, respectively. 
Unlike previous MLLMs, in which only answer token $Y_a$ generated loss, we attempted to align all token logits between the teacher and student models. We observe that the performance will decrease for all aspects we have tested. 

\begin{table*}[tp]
    \centering
    \setlength{\tabcolsep}{4mm}{\begin{tabular}{c|cccc}
    \toprule
    Method & VQA$^\text{T}$ & SQA$^\text{I}$ & POPE & MME$^\text{P}$ \\
    \midrule
    \rowcolor{whitesmoke} Baseline & 47.5 & 59.7 & 85.2 & 1307.7 \\
    MSE & 35.9 & 51.8 & 83.3 & 1159.8 \\
    Forward KL  & 48.2 &  57.2  & 86.0 &  1313.3 \\
    Reverse KL~\cite{minillm} & 48.6 & 56.1 & 85.7 & 1252.1 \\
    JSD ~\cite{wen2023fdivergence} & 48.7 & 59.3  & 85.9 & 1301.2 \\
    Forward KL + Logit stand.~\cite{Sun2024Logit} & 48.6 & 58.4 & 86.1 & 1254.4 \\
    Forward KL + All tokens  & 45.9 & 56.2 & 85.4 & 1294.1 \\
    \bottomrule
    \end{tabular}}
    \caption{Knowledge distillation for aligning logits using various methods. Logit stand. represents logit standardization.}
    \label{tab:logit}
\end{table*}

\begin{tcolorbox}[
colframe=black,
boxsep=1pt,
]
    \paragraph{\textbf{\text{Finding} 2.}} Aligning the classification logits of answer tokens using KL divergence is effective. Different KL methods achieve comparable results.
\end{tcolorbox}

\subsection{Affinity aware distillation}
As previously described, the MLLMs are trained using image-text pairs or image-based instruction tuning datasets. These models are usually required to describe images or answer questions about images. Consequently, a well-trained MLLM captures intricate relationships between visual and textual information, generating coherent and contextually relevant descriptions or responses. As previous methods~\cite{yang2023clipkd,tinyclip,Wang2022CLIPTDCT} have explored the image-text affinity distillation for the CLIP model, they have demonstrated the effectiveness of leveraging the relationship between images and corresponding textual descriptions to improve model performance. We want to test whether affinity-aware distillation is effective for MLLMs. To this end, we design two methods to explore the affinity for MLLMs. The first method involves aligning the attention score between the teacher and student model using MSE loss. We also consider the attention score of various token types, including the attention score between image and answer tokens. The second method involves aligning the feature embeddings' similarity between the teacher and student models. Specifically, we calculate the cosine similarity of image embeddings $H_v$ and text embeddings $H_q$. The results show that either the attention score or the 
cosine similarity does not perform well. 
Affinity distillation is well-suited for CLIP-like models as the compatibility between visual and text features aligns better with contrastive loss. 
The next-token prediction paradigm is not suitable for affinity supervision.

\begin{table*}[tp]
    \centering
     \setlength{\tabcolsep}{4mm}{\begin{tabular}{c|cccc}
    \toprule
    Method & VQA$^\text{T}$ & SQA$^\text{I}$ & POPE & MME$^\text{P}$ \\
    \midrule
    \rowcolor{whitesmoke} Baseline & 47.5 & 59.7 & 85.2 & 1307.7 \\
    Attention Score & 46.2 & 58.1 & 82.3 & 1279.8 \\
    Similarity~\cite{yang2023clipkd} & 44.7 & 52.0 &  82.4 & 1120.3 \\
    \toprule
    \end{tabular}
    }
    \caption{Affinity aware distillation.}
    \label{tab:affinity}
\end{table*}

\begin{tcolorbox}[
colframe=black,
arc=4pt,
boxsep=1pt,
fonttitle=\bfseries\itshape,
]
    \paragraph{\textbf{\text{Finding} 3.}} Affinity-aware distillation is not well-suited for MLLMs. Affinity supervision is more suitable for contrastive loss than for auto-regressive loss.
\end{tcolorbox}

\subsection{Data-driven knowledge distillation}
Our experiment uses the 665K mixed dataset from LLaVA~\cite{liu2023llava} as the fine-tuning dataset. However, 150K instruction-tuning datasets among them are generated by the GPT-3.5, which is a language-only model. This may introduce limitations in visual understanding tasks, as the instruction-tuning data generated by GPT-3.5 lacks visual context, potentially impacting the performance of models trained on this mixed dataset in tasks requiring nuanced visual comprehension.  
There also exists a distribution gap between the instruction-tuning dataset and the initial large language model, which can result in decreased performance, as discussed in ~\cite{sdft}. To explore data-driven knowledge distillation, we regenerate the 150K instruction-tuning datasets. In the first row of Tab.~\ref{tab:data_driven}, we keep the original images and questions unchanged while regenerating the answer for each question using the teacher model. The results show that data corrected by the teacher model has performed well in VQA$^\text{T}$, SQA$^\text{I}$, and POPE, but has slightly declined in MME$^\text{P}$. 

The data produced by the teacher model may have distribution discrepancies compared to the student model. The student model's limited capacity makes it challenging to replicate the teacher model's data distribution completely. In the second row of Tab.~\ref{tab:data_driven}, we try to generate data from the student model and let the more powerful teacher model mimic the distribution of the student model. Specifically, we use the student model being trained to generate data and then use this data as supervision distillation. Because data is generated serially, the training process takes too long. To address this issue, we randomly regenerate 50\% of the data to obtain the answer. This method only demonstrates comparable performance and even performs worse on this SQA$^\text{I}$. However, we believe there is still room for further exploration of this method, such as optimizing student models more efficiently and efficiently, rather than randomly selecting data.

To test the scalability of the method, we also try to expand the instruction-tuning dataset from 665K to 2M~\cite{chu2024mobilevlmv2}, which adds 205K VQA dataset, 1.4M caption dataset, and 123K conversation dataset. In the third row of Tab.~\ref{tab:data_driven}, we present the baseline performance trained using the 2M instructing-tuning dataset. 
The fourth row of Tab.~\ref{tab:data_driven} shows a slight improvement. We can observe that the results trained using the more complex instruction-tuning dataset show a much greater improvement in model performance. 

\begin{table}[tp]
    \centering
    \setlength{\tabcolsep}{4mm}{\begin{tabular}{c|cccc}
    \toprule
    Method & VQA$^\text{T}$ & SQA$^\text{I}$ & POPE & MME$^\text{P}$ \\
    \midrule
    Data from teacher & 49.2 & 60.7 & 85.4 & 1296.5 \\
    Data from student & 49.8 & 57.9 & 85.9 & 1276.3 \\
    Baseline+2M & 57.4 & 66.9 & 85.5 & 1386.8 \\
    Distillation + 2M & 58.2 & 68.4 & 86.0 & 1365.9 \\
    \toprule
    \end{tabular}
    }
    \caption{Data-driven knowledge distillation.}
    \label{tab:data_driven}
\end{table}

\begin{tcolorbox}[
colframe=black,
arc=4pt,
boxsep=1pt,
fonttitle=\bfseries\itshape,
]
    \paragraph{\textbf{\text{Finding} 4.}} Instruction-tuning data generated by the teacher model or the student model can boost the performance well.
\end{tcolorbox}

\subsection{Compared with other distillation strategy}
DistilBERT~\cite{sanh2019distilbert} proposes an effective and efficient distillation that reduces the size of a BERT model by 40\% while retaining 97\% of its language understanding capabilities and being 60\% faster. 
It shares the overall architecture of BERT~\cite{devlin2018bert}, but with half the number of layers. The student model is initialized from the teacher model by removing every other layer. The training objective of DistilBERT consists of a \textit{soft target probabilities} supervision, \textit{masked language modeling} loss, and \textit{cosine embedding} loss. We have also implemented this strategy on MLLMs.  We replace masked language modeling loss with autoregressive loss. We modify the cosine embeddings loss and soft target probabilities supervision to only be applied to the answer tokens $Y_a$ of the last transformers layer. The student model has half the number of layers as the teacher model LLaVA-v1.5-13B, and its parameters are initialized by taking every other layer. 
In Tab.~\ref{tab:distilbert}, we can observe that although this method has almost twice the number of parameters as ours, its performance is not as good as ours.

\begin{table}[t]
    \centering
    \setlength{\tabcolsep}{4mm}{\begin{tabular}{c|cccc}
    \toprule
    Method & VQA$^\text{T}$ & SQA$^\text{I}$ & POPE & MME$^\text{P}$ \\
    \midrule
     DistilBERT~\cite{sanh2019distilbert} & 46.5 & 55.3 & 85.4 & 1296.0 \\
    {\ourmethod}   & 50.7 & 64.1 & 86.7 & 1376.1 \\
    \bottomrule
    \end{tabular}
    \caption{Compared with distillation strategy from DistilBERT. }
    }
    \label{tab:distilbert}
\end{table}

\subsection{Performance evaluation}
In Table \ref{tab:main_table}, we conduct a comprehensive comparison with current MLLMs. We utilize logit-level, feature-level distillation, and data-driven knowledge distillation techniques. For logit-level alignment, we apply forward KL without norm standardization. In feature-level distillation, we match all token representations of the final layer between the teacher and student models. As for data-driven knowledge distillation, we recreate the original 150K instruction tuning dataset using the teacher model to ensure similarity. Our results are based on training models using 1.4B and 2.7B large language models combined with the CLIP ViT-L/14 visual encoder, respectively. Regardless of whether we use the 1.4B or 2.7B language models, our distilled strategy-trained models outperform the baseline model. The performance on VQA$^\text{T}$ of our method is even comparable with InstructBLIP-13B~\cite{dai2023instructblip}. Compared to MoE-LLaVA-2.7B~\cite{lin2024moe}, which utilizes the more potent Phi-2.7~\cite{zhu2024llavaphi} language model, our distillation approach yields slightly superior results on GQA, SQA$^\text{I}$, VQA$^\text{T}$, and POPE. This showcases that our distillation strategy can outperform even more powerful MOE methods.



\begin{table*}[tp]
\centering
\setlength{\tabcolsep}{4pt}
\small
\scalebox{0.90}{
\begin{tabular}{*{1}{l}*{2}{l}|*{7}{c}}
\toprule
Method & LLM & Res. & GQA & SQA$^\text{I}$ & VQA$^\text{T}$ & POPE & MME$^\text{P}$ & MMB$^\text{dev}$ & Avg.\\
\midrule
BLIP-2 \cite{li2023blip2} & Vicuna-13B & 224  & 41.0 & 61.0 & 42.5 & 85.3 & 1293.8 & -- & -  \\
InstructBLIP \cite{dai2023instructblip}& Vicuna-13B & 224 & 49.5  & 63.1 & 50.7 & 78.9 & 1212.8 & -- & -  \\
Shikra \cite{chen2023shikra}& Vicuna-13B & 224  & -- & -- & -- & -- & -- & 58.8 & --  \\

Openflamingo \cite{anas2023OpenFlamingo} & MPT-7B & 336 & -- & -- & 33.6 & -- & -- & 4.6 & -- \\
Qwen-VL \cite{Qwen-VL} & Qwen-7B & 448& 59.3 & 67.1 & 63.8 & -- & 1487.6 & 38.2 & - \\

IDEFICS-9B \cite{laurencon2023obelics} & LLaMA-7B & 224& 38.4 & -- & 25.9 & -- & -- & 48.2 & -- \\

MiniGPT-v2 \cite{chen2023minigpt} & LLaMA-7B & 448 & 60.3 & -- & -- & -- & -- & 12.2 & --\\
InstructBLIP \cite{dai2023instructblip} & Vicuna-7B & 224  & 49.2  & 60.5 & 50.1 & -- & -- & 36.0 & -- \\
LLaVA-1.5 \cite{liu2023improved} & Vicuna-7B & 336  & 62.0 & 66.8 & 58.2 & 85.9 & 1510.7 & 64.3 & 68.8 \\
ShareGPT4V \cite{chen2023sharegpt4v} & Vicuna-7B & 336  & 63.3 & 68.4 & 60.4 & 85.7 & 1567.4 & 68.8 & 70.8 \\
\hline
MoE-LLaVA-1.6B$\times$4 \cite{lin2024moe} & StableLM-1.6B& 336 & 60.4& 62.6 & 47.8 &84.3 & 1300.8$^\text{*}$ & 59.4 & 63.3 \\
MoE-LLaVA-2.7B$\times$4 \cite{lin2024moe} & Phi-2.7B& 336 &  61.1& 68.7 & 50.2 &85.0 & 1396.4$^\text{*}$ & 65.5 & 66.7 \\
MobileVLM 1.7B \cite{chu2023mobilevlm} & MobileLLaMA 1.4B & 336& 56.1 & 57.3 & 41.5 & 84.5 & 1196.2 & 53.2 & 58.7 \\
MobileVLM 3B \cite{chu2023mobilevlm} & MobileLLaMA 2.7B & 336  & 59.0 & 61.2 & 47.5 & 84.9 & 1288.9 & 59.6 & 62.8 \\
\midrule
{\ourmethod} & MobileLLaMA 1.4B & 336 & 55.4 & 56.0 & 45.3 & 84.7 & 1178.6 & 55.0 & 59.2 \\ 
{\ourmethod} & MobileLLaMA 2.7B  & 336 & 61.4 &  64.1 & 50.7 & 86.7 & 1376.1 & 62.5 & 65.7\\
\bottomrule
\end{tabular}
}
\caption{\textbf{Comparison with SOTA methods on six VLM benchmarks.}  Column \textit{Res.} is the image resolution of the vision model. Column \textit{Avg.} indicates the average accuracy on six evaluation benchmarks. The values in the MME$^\text{P}$ column should be divided by 2000 before being included in the average accuracy calculation.}
\label{tab:main_table}
\end{table*}

\section{Conclusion}
\label{sec:conclusion}

In this study, we conducted extensive experiments to investigate the key factors in distilling multimodal large language models. Our results show that aligning the classification logits and hidden embeddings of the final layer is effective. We also confirmed the effectiveness of data-driven knowledge distillation by generating answers using either teacher or student models. Additionally, when trained with a 2M instruction-tuning dataset, our distillation strategies performed well, demonstrating their scalability. We hope our findings will help guide future research on distilling multimodal models and inspire further advancements in transferring capabilities from large to small models.

\clearpage

{\small
	\bibliographystyle{ieee_fullname}
	\bibliography{refbib}

\begin{thebibliography}{10}\itemsep=-1pt

\bibitem{agarwal2023onpolicy}
Rishabh Agarwal, Nino Vieillard, Yongchao Zhou, Piotr Stanczyk, Sabela Ramos,
  Matthieu Geist, and Olivier Bachem.
\newblock On-policy distillation of language models: Learning from
  self-generated mistakes.
\newblock In {\em ICLR}, 2024.

\bibitem{Agarwal2023OnPolicyDO}
Rishabh Agarwal, Nino Vieillard, Yongchao Zhou, Piotr Stańczyk, Sabela Ramos,
  Matthieu Geist, and Olivier Bachem.
\newblock On-policy distillation of language models: Learning from
  self-generated mistakes.
\newblock In {\em ICLR}, 2023.

\bibitem{Alayrac2022FlamingoAV}
Jean-Baptiste Alayrac, Jeff Donahue, Pauline Luc, Antoine Miech, Iain Barr,
  Yana Hasson, Karel Lenc, Arthur Mensch, Katie Millican, Malcolm Reynolds,
  Roman Ring, Eliza Rutherford, Serkan Cabi, Tengda Han, Zhitao Gong, Sina
  Samangooei, Marianne Monteiro, Jacob Menick, Sebastian Borgeaud, Andy Brock,
  Aida Nematzadeh, Sahand Sharifzadeh, Mikolaj Binkowski, Ricardo Barreira,
  Oriol Vinyals, Andrew Zisserman, and Karen Simonyan.
\newblock Flamingo: a visual language model for few-shot learning.
\newblock In {\em NeurIPS}, 2022.

\bibitem{awadalla2023openflamingo}
Anas Awadalla, Irena Gao, Josh Gardner, Jack Hessel, Yusuf Hanafy, Wanrong Zhu,
  Kalyani Marathe, Yonatan Bitton, Samir Gadre, Shiori Sagawa, Jenia Jitsev,
  Simon Kornblith, Pang~Wei Koh, Gabriel Ilharco, Mitchell Wortsman, and Ludwig
  Schmidt.
\newblock Openflamingo: An open-source framework for training large
  autoregressive vision-language models.
\newblock {\em arXiv preprint arXiv:2308.01390}, 2023.

\bibitem{anas2023OpenFlamingo}
Anas Awadalla, Irena Gao, Josh Gardner, Jack Hessel, Yusuf Hanafy, Wanrong Zhu,
  Kalyani Marathe, Yonatan Bitton, Samir Gadre, Shiori Sagawa, Jenia Jitsev,
  Simon Kornblith, Pang~Wei Koh, Gabriel Ilharco, Mitchell Wortsman, and Ludwig
  Schmidt.
\newblock Openflamingo: An open-source framework for training large
  autoregressive vision-language models.
\newblock {\em arXiv preprint arXiv:2308.01390}, 2023.

\bibitem{Qwen-VL}
Jinze Bai, Shuai Bai, Shusheng Yang, Shijie Wang, Sinan Tan, Peng Wang, Junyang
  Lin, Chang Zhou, and Jingren Zhou.
\newblock Qwen-vl: A versatile vision-language model for understanding,
  localization, text reading, and beyond.
\newblock {\em arXiv preprint arXiv:2308.12966}, 2023.

\bibitem{brown2020language}
Tom Brown, Benjamin Mann, Nick Ryder, Melanie Subbiah, Jared~D Kaplan, Prafulla
  Dhariwal, Arvind Neelakantan, Pranav Shyam, Girish Sastry, Amanda Askell,
  et~al.
\newblock Language models are few-shot learners.
\newblock In {\em NeurIPS}, 2020.

\bibitem{cai2024internlm2}
Zheng Cai, Maosong Cao, Haojiong Chen, Kai Chen, Keyu Chen, Xin Chen, Xun Chen,
  Zehui Chen, Zhi Chen, Pei Chu, Xiaoyi Dong, Haodong Duan, Qi Fan, Zhaoye Fei,
  Yang Gao, Jiaye Ge, Chenya Gu, Yuzhe Gu, Tao Gui, Aijia Guo, Qipeng Guo,
  Conghui He, Yingfan Hu, Ting Huang, Tao Jiang, Penglong Jiao, Zhenjiang Jin,
  Zhikai Lei, Jiaxing Li, Jingwen Li, Linyang Li, Shuaibin Li, Wei Li, Yining
  Li, Hongwei Liu, Jiangning Liu, Jiawei Hong, Kaiwen Liu, Kuikun Liu, Xiaoran
  Liu, Chengqi Lv, Haijun Lv, Kai Lv, Li Ma, Runyuan Ma, Zerun Ma, Wenchang
  Ning, Linke Ouyang, Jiantao Qiu, Yuan Qu, Fukai Shang, Yunfan Shao, Demin
  Song, Zifan Song, Zhihao Sui, Peng Sun, Yu Sun, Huanze Tang, Bin Wang,
  Guoteng Wang, Jiaqi Wang, Jiayu Wang, Rui Wang, Yudong Wang, Ziyi Wang,
  Xingjian Wei, Qizhen Weng, Fan Wu, Yingtong Xiong, Chao Xu, Ruiliang Xu, Hang
  Yan, Yirong Yan, Xiaogui Yang, Haochen Ye, Huaiyuan Ying, Jia Yu, Jing Yu,
  Yuhang Zang, Chuyu Zhang, Li Zhang, Pan Zhang, Peng Zhang, Ruijie Zhang, Shuo
  Zhang, Songyang Zhang, Wenjian Zhang, Wenwei Zhang, Xingcheng Zhang, Xinyue
  Zhang, Hui Zhao, Qian Zhao, Xiaomeng Zhao, Fengzhe Zhou, Zaida Zhou, Jingming
  Zhuo, Yicheng Zou, Xipeng Qiu, Yu Qiao, and Dahua Lin.
\newblock Internlm2 technical report, 2024.

\bibitem{chen2023minigpt}
Jun Chen, Deyao Zhu, Xiaoqian Shen, Xiang Li, Zechun Liu, Pengchuan Zhang,
  Raghuraman Krishnamoorthi, Vikas Chandra, Yunyang Xiong, and Mohamed
  Elhoseiny.
\newblock Minigpt-v2: large language model as a unified interface for
  vision-language multi-task learning.
\newblock {\em arXiv preprint arXiv:2310.09478}, 2023.

\bibitem{chen2023shikra}
Keqin Chen, Zhao Zhang, Weili Zeng, Richong Zhang, Feng Zhu, and Rui Zhao.
\newblock Shikra: Unleashing multimodal llm's referential dialogue magic.
\newblock {\em arXiv preprint arXiv:2306.15195}, 2023.

\bibitem{chen2023sharegpt4v}
Lin Chen, Jisong Li, Xiaoyi Dong, Pan Zhang, Conghui He, Jiaqi Wang, Feng Zhao,
  and Dahua Lin.
\newblock Sharegpt4v: Improving large multi-modal models with better captions.
\newblock {\em arXiv preprint arXiv:2311.12793}, 2023.

\bibitem{chen2024image}
Liang Chen, Haozhe Zhao, Tianyu Liu, Shuai Bai, Junyang Lin, Chang Zhou, and
  Baobao Chang.
\newblock An image is worth 1/2 tokens after layer 2: Plug-and-play inference
  acceleration for large vision-language models.
\newblock {\em arXiv:2403.06764}, 2024.

\bibitem{Chen2015MicrosoftCC}
Xinlei Chen, Hao Fang, Tsung-Yi Lin, Ramakrishna Vedantam, Saurabh Gupta, Piotr
  Doll{\'a}r, and C.~Lawrence Zitnick.
\newblock Microsoft coco captions: Data collection and evaluation server.
\newblock {\em ArXiv}, abs/1504.00325, 2015.

\bibitem{vicuna2023}
Wei-Lin Chiang, Zhuohan Li, Zi Lin, Ying Sheng, Zhanghao Wu, Hao Zhang, Lianmin
  Zheng, Siyuan Zhuang, Yonghao Zhuang, Joseph~E. Gonzalez, Ion Stoica, and
  Eric~P. Xing.
\newblock Vicuna: An open-source chatbot impressing gpt-4 with 90\%* chatgpt
  quality, March 2023.

\bibitem{chu2023mobilevlm}
Xiangxiang Chu, Limeng Qiao, Xinyang Lin, Shuang Xu, Yang Yang, Yiming Hu, Fei
  Wei, Xinyu Zhang, Bo Zhang, Xiaolin Wei, et~al.
\newblock Mobilevlm: A fast, reproducible and strong vision language assistant
  for mobile devices.
\newblock {\em arXiv preprint arXiv:2312.16886}, 2023.

\bibitem{chu2024mobilevlmv2}
Xiangxiang Chu, Limeng Qiao, Xinyu Zhang, Shuang Xu, Fei Wei, Yang Yang,
  Xiaofei Sun, Yiming Hu, Xinyang Lin, Bo Zhang, and Chunhua Shen.
\newblock Mobilevlm v2: Faster and stronger baseline for vision language model,
  2024.

\bibitem{dai2023instructblip}
Wenliang Dai, Junnan Li, Dongxu Li, Anthony Meng~Huat Tiong, Junqi Zhao,
  Weisheng Wang, Boyang Li, Pascale Fung, and Steven Hoi.
\newblock Instructblip: Towards general-purpose vision-language models with
  instruction tuning, 2023.

\bibitem{Das_2017}
Abhishek Das, Satwik Kottur, Khushi Gupta, Avi Singh, Deshraj Yadav, Jose M.~F.
  Moura, Devi Parikh, and Dhruv Batra.
\newblock Visual dialog.
\newblock In {\em CVPR}, 2017.

\bibitem{devlin2018bert}
Jacob Devlin, Ming-Wei Chang, Kenton Lee, and Kristina Toutanova.
\newblock Bert: Pre-training of deep bidirectional transformers for language
  understanding.
\newblock In {\em NAACL}, 2019.

\bibitem{ding2023enhancing}
Ning Ding, Yulin Chen, Bokai Xu, Yujia Qin, Zhi Zheng, Shengding Hu, Zhiyuan
  Liu, Maosong Sun, and Bowen Zhou.
\newblock Enhancing chat language models by scaling high-quality instructional
  conversations.
\newblock In {\em EMNLP}, 2023.

\bibitem{fang2021compressing}
Zhiyuan Fang, Jianfeng Wang, Xiaowei Hu, Lijuan Wang, Yezhou Yang, and Zicheng
  Liu.
\newblock Compressing visual-linguistic model via knowledge distillation.
\newblock In {\em ICCV}, 2021.

\bibitem{fu2023mme}
Chaoyou Fu, Peixian Chen, Yunhang Shen, Yulei Qin, Mengdan Zhang, Xu Lin,
  Jinrui Yang, Xiawu Zheng, Ke Li, Xing Sun, Yunsheng Wu, and Rongrong Ji.
\newblock Mme: A comprehensive evaluation benchmark for multimodal large
  language models.
\newblock {\em arXiv preprint arXiv:2306.13394}, 2023.

\bibitem{minillm}
Yuxian Gu, Li Dong, Furu Wei, and Minlie Huang.
\newblock Knowledge distillation of large language models.
\newblock In {\em ICLR}, 2024.

\bibitem{guan2020differentiable}
Yushuo Guan, Pengyu Zhao, Bingxuan Wang, Yuanxing Zhang, Cong Yao, Kaigui Bian,
  and Jian Tang.
\newblock Differentiable feature aggregation search for knowledge distillation.
\newblock In {\em ECCV}, 2020.

\bibitem{hinton2015distilling}
Geoffrey Hinton, Oriol Vinyals, and Jeff Dean.
\newblock Distilling the knowledge in a neural network, 2015.

\bibitem{Hou2020DynaBERTDB}
Lu Hou, Zhiqi Huang, Lifeng Shang, Xin Jiang, and Qun Liu.
\newblock Dynabert: Dynamic bert with adaptive width and depth.
\newblock {\em NeurIPS}, 2020.

\bibitem{hudson2019gqa}
Drew~A Hudson and Christopher~D Manning.
\newblock Gqa: A new dataset for real-world visual reasoning and compositional
  question answering.
\newblock In {\em CVPR}, 2019.

\bibitem{tinybert}
Xiaoqi Jiao, Yichun Yin, Lifeng Shang, Xin Jiang, Xiao Chen, Linlin Li, Fang
  Wang, and Qun Liu.
\newblock Tinybert: Distilling bert for natural language understanding.
\newblock In {\em EMNLP}, 2020.

\bibitem{ko2024distillm}
Jongwoo Ko, Sungnyun Kim, Tianyi Chen, and Se-Young Yun.
\newblock Distillm: Towards streamlined distillation for large language models.
\newblock {\em arXiv preprint arXiv:2402.03898}, 2024.

\bibitem{laurencon2023obelics}
Hugo Laurençon, Lucile Saulnier, Léo Tronchon, Stas Bekman, Amanpreet Singh,
  Anton Lozhkov, Thomas Wang, Siddharth Karamcheti, Alexander~M. Rush, Douwe
  Kiela, Matthieu Cord, and Victor Sanh.
\newblock Obelics: An open web-scale filtered dataset of interleaved image-text
  documents, 2023.

\bibitem{li2023blip2}
Junnan Li, Dongxu Li, Silvio Savarese, and Steven Hoi.
\newblock Blip-2: Bootstrapping language-image pre-training with frozen image
  encoders and large language models, 2023.

\bibitem{li2022blip}
Junnan Li, Dongxu Li, Caiming Xiong, and Steven Hoi.
\newblock Blip: Bootstrapping language-image pre-training for unified
  vision-language understanding and generation, 2022.

\bibitem{li2023evaluating}
Yifan Li, Yifan Du, Kun Zhou, Jinpeng Wang, Wayne~Xin Zhao, and Ji-Rong Wen.
\newblock Evaluating object hallucination in large vision-language models.
\newblock {\em EMNLP}, 2023.

\bibitem{liang2022taskaware}
Chen Liang, Simiao Zuo, Qingru Zhang, Pengcheng He, Weizhu Chen, and Tuo Zhao.
\newblock Less is more: Task-aware layer-wise distillation for language model
  compression.
\newblock In {\em ICML}, 2023.

\bibitem{liang2022less}
Chen Liang, Simiao Zuo, Qingru Zhang, Pengcheng He, Weizhu Chen, and Tuo Zhao.
\newblock Less is more: Task-aware layer-wise distillation for language model
  compression.
\newblock {\em ICML}, 2023.

\bibitem{lin2024moe}
Bin Lin, Zhenyu Tang, Yang Ye, Jiaxi Cui, Bin Zhu, Peng Jin, Junwu Zhang, Munan
  Ning, and Li Yuan.
\newblock Moe-llava: Mixture of experts for large vision-language models.
\newblock {\em arXiv preprint arXiv:2401.15947}, 2024.

\bibitem{lin2024boosting}
Zhihang Lin, Mingbao Lin, Luxi Lin, and Rongrong Ji.
\newblock Boosting multimodal large language models with visual tokens
  withdrawal for rapid inference.
\newblock {\em arXiv:2405.05803}, 2024.

\bibitem{mftcoder2023}
Bingchang Liu, Chaoyu Chen, Cong Liao, Zi Gong, Huan Wang, Zhichao Lei, Ming
  Liang, Dajun Chen, Min Shen, Hailian Zhou, Hang Yu, and Jianguo Li.
\newblock Mftcoder: Boosting code llms with multitask fine-tuning.
\newblock {\em arXiv preprint arXiv}, 2023.

\bibitem{Liu_2023}
Fangyu Liu, Guy Emerson, and Nigel Collier.
\newblock Visual spatial reasoning.
\newblock {\em ACL}, 2023.

\bibitem{liu2023improved}
Haotian Liu, Chunyuan Li, Yuheng Li, and Yong~Jae Lee.
\newblock Improved baselines with visual instruction tuning.
\newblock {\em CVPR}, 2024.

\bibitem{liu2023llava}
Haotian Liu, Chunyuan Li, Qingyang Wu, and Yong~Jae Lee.
\newblock Visual instruction tuning.
\newblock In {\em NeurIPS}, 2023.

\bibitem{liu2019structured}
Yifan Liu, Ke Chen, Chris Liu, Zengchang Qin, Zhenbo Luo, and Jingdong Wang.
\newblock Structured knowledge distillation for semantic segmentation.
\newblock In {\em CVPR}, 2019.

\bibitem{liu2023mmbench}
Yuan Liu, Haodong Duan, Yuanhan Zhang, Bo Li, Songyang Zhang, Wangbo Zhao, Yike
  Yuan, Jiaqi Wang, Conghui He, Ziwei Liu, et~al.
\newblock Mmbench: Is your multi-modal model an all-around player?
\newblock {\em arXiv preprint arXiv:2307.06281}, 2023.

\bibitem{Loshchilov2017FixingWD}
Ilya Loshchilov and Frank Hutter.
\newblock Fixing weight decay regularization in adam.
\newblock {\em ICLR}, 2019.

\bibitem{lu2022learn}
Pan Lu, Swaroop Mishra, Tony Xia, Liang Qiu, Kai-Wei Chang, Song-Chun Zhu,
  Oyvind Tafjord, Peter Clark, and Ashwin Kalyan.
\newblock Learn to explain: Multimodal reasoning via thought chains for science
  question answering.
\newblock In {\em NeurIPS}, 2022.

\bibitem{Lu2022LearnTE}
Pan Lu, Swaroop Mishra, Tony Xia, Liang Qiu, Kai-Wei Chang, Song-Chun Zhu,
  Oyvind Tafjord, Peter Clark, and A. Kalyan.
\newblock Learn to explain: Multimodal reasoning via thought chains for science
  question answering.
\newblock {\em ArXiv}, abs/2209.09513, 2022.

\bibitem{Lu2021IconQAAN}
Pan Lu, Liang Qiu, Jiaqi Chen, Tony Xia, Yizhou Zhao, Wei Zhang, Zhou Yu,
  Xiaodan Liang, and Song-Chun Zhu.
\newblock Iconqa: A new benchmark for abstract diagram understanding and visual
  language reasoning.
\newblock {\em ArXiv}, abs/2110.13214, 2021.

\bibitem{ChatGPT}
OpenAI.
\newblock Chatgpt.
\newblock \url{https://openai.com/blog/chatgpt/}, 2023.

\bibitem{GPT4}
OpenAI.
\newblock Gpt-4 technical report.
\newblock {\em arXiv:2303.08774}, 2023.

\bibitem{ordonez2011im2text}
Vicente Ordonez, Girish Kulkarni, and Tamara Berg.
\newblock Im2text: Describing images using 1 million captioned photographs.
\newblock {\em Advances in neural information processing systems}, 24, 2011.

\bibitem{qi2024generalizable}
Lu Qi, Yi-Wen Chen, Lehan Yang, Tiancheng Shen, Xiangtai Li, Weidong Guo, Yu
  Xu, and Ming-Hsuan Yang.
\newblock Generalizable entity grounding via assistance of large language
  model.
\newblock {\em arXiv preprint arXiv:2402.02555}, 2024.

\bibitem{CLIP}
Alec Radford, Jong~Wook Kim, Chris Hallacy, Aditya Ramesh, Gabriel Goh,
  Sandhini Agarwal, Girish Sastry, Amanda Askell, Pamela Mishkin, Jack Clark,
  et~al.
\newblock Learning transferable visual models from natural language
  supervision.
\newblock In {\em ICML}, 2021.

\bibitem{hanoona2023GLaMM}
Hanoona Rasheed, Muhammad Maaz, Sahal Shaji, Abdelrahman Shaker, Salman Khan,
  Hisham Cholakkal, Rao~M. Anwer, Eric Xing, Ming-Hsuan Yang, and Fahad~S.
  Khan.
\newblock Glamm: Pixel grounding large multimodal model.
\newblock {\em CVPR}, 2024.

\bibitem{sanh2019distilbert}
Victor Sanh, Lysandre Debut, Julien Chaumond, and Thomas Wolf.
\newblock Distilbert, a distilled version of bert: smaller, faster, cheaper and
  lighter, 2019.

\bibitem{shu2021channel}
Changyong Shu, Yifan Liu, Jianfei Gao, Zheng Yan, and Chunhua Shen.
\newblock Channel-wise knowledge distillation for dense prediction.
\newblock In {\em ICCV}, 2021.

\bibitem{Singh_2019}
Amanpreet Singh, Vivek Natarajan, Meet Shah, Yu Jiang, Xinlei Chen, Dhruv
  Batra, Devi Parikh, and Marcus Rohrbach.
\newblock Towards vqa models that can read.
\newblock In {\em CVPR}, 2019.

\bibitem{singh2019towards}
Amanpreet Singh, Vivek Natarjan, Meet Shah, Yu Jiang, Xinlei Chen, Devi Parikh,
  and Marcus Rohrbach.
\newblock Towards vqa models that can read.
\newblock In {\em CVPR}, 2019.

\bibitem{Sun2019PatientKD}
S. Sun, Yu Cheng, Zhe Gan, and Jingjing Liu.
\newblock Patient knowledge distillation for bert model compression.
\newblock In {\em EMNLP}, 2019.

\bibitem{Sun2024Logit}
Shangquan Sun, Wenqi Ren, Jingzhi Li, Rui Wang, and Xiaochun Cao.
\newblock Logit standardization in knowledge distillation.
\newblock In {\em CVPR}, 2024.

\bibitem{Sun2020MobileBERTAC}
Zhiqing Sun, Hongkun Yu, Xiaodan Song, Renjie Liu, Yiming Yang, and Denny Zhou.
\newblock Mobilebert: a compact task-agnostic bert for resource-limited
  devices.
\newblock In {\em ACL}, 2020.

\bibitem{alpaca}
Rohan Taori, Ishaan Gulrajani, Tianyi Zhang, Yann Dubois, Xuechen Li, Carlos
  Guestrin, Percy Liang, and Tatsunori~B. Hashimoto.
\newblock Stanford alpaca: An instruction-following llama model.
\newblock \url{https://github.com/tatsu-lab/stanford_alpaca}, 2023.

\bibitem{team2023gemini}
Gemini Team, Rohan Anil, Sebastian Borgeaud, Yonghui Wu, Jean-Baptiste Alayrac,
  Jiahui Yu, Radu Soricut, Johan Schalkwyk, Andrew~M Dai, Anja Hauth, et~al.
\newblock Gemini: a family of highly capable multimodal models.
\newblock {\em arXiv preprint arXiv:2312.11805}, 2023.

\bibitem{Achiam2023GPT4TR}
OpenAI teams.
\newblock Gpt-4 technical report.
\newblock {\em Arxiv}, 2023.

\bibitem{touvron2023llama}
Hugo Touvron, Thibaut Lavril, Gautier Izacard, Xavier Martinet, Marie-Anne
  Lachaux, Timothée Lacroix, Baptiste Rozière, Naman Goyal, Eric Hambro,
  Faisal Azhar, Aurelien Rodriguez, Armand Joulin, Edouard Grave, and Guillaume
  Lample.
\newblock Llama: Open and efficient foundation language models, 2023.

\bibitem{Vaswani2017AttentionIA}
Ashish Vaswani, Noam~M. Shazeer, Niki Parmar, Jakob Uszkoreit, Llion Jones,
  Aidan~N. Gomez, Lukasz Kaiser, and Illia Polosukhin.
\newblock Attention is all you need.
\newblock In {\em NeurIPS}, 2017.

\bibitem{Wang2023VIGCVI}
Bin Wang, Fan Wu, Xiao Han, Jiahui Peng, Huaping Zhong, Pan Zhang, Xiao wen
  Dong, Weijia Li, Wei Li, Jiaqi Wang, and Conghui He.
\newblock Vigc: Visual instruction generation and correction.
\newblock {\em arXiv:2308.12714}, 2023.

\bibitem{wang2019distilling}
Tao Wang, Li Yuan, Xiaopeng Zhang, and Jiashi Feng.
\newblock Distilling object detectors with fine-grained feature imitation.
\newblock In {\em CVPR}, 2019.

\bibitem{wang2022selfinstruct}
Yizhong Wang, Yeganeh Kordi, Swaroop Mishra, Alisa Liu, Noah~A. Smith, Daniel
  Khashabi, and Hannaneh Hajishirzi.
\newblock Self-instruct: Aligning language models with self-generated
  instructions.
\newblock In {\em ACL}, 2023.

\bibitem{wang2020intra}
Yukang Wang, Wei Zhou, Tao Jiang, Xiang Bai, and Yongchao Xu.
\newblock Intra-class feature variation distillation for semantic segmentation.
\newblock In {\em ECCV}, 2020.

\bibitem{wang2022multimodal}
Zhecan Wang, Noel Codella, Yen-Chun Chen, Luowei Zhou, Xiyang Dai, Bin Xiao,
  Jianwei Yang, Haoxuan You, Kai-Wei Chang, Shih-fu Chang, et~al.
\newblock Multimodal adaptive distillation for leveraging unimodal encoders for
  vision-language tasks.
\newblock {\em arXiv preprint arXiv:2204.10496}, 2022.

\bibitem{Wang2022CLIPTDCT}
Zhecan Wang, Noel C.~F. Codella, Yen-Chun Chen, Luowei Zhou, Jianwei Yang,
  Xiyang Dai, Bin Xiao, Haoxuan You, Shih-Fu Chang, and Lu Yuan.
\newblock Clip-td: Clip targeted distillation for vision-language tasks.
\newblock {\em arXiv:2201.05729}, 2022.

\bibitem{wen2023fdivergence}
Yuqiao Wen, Zichao Li, Wenyu Du, and Lili Mou.
\newblock f-divergence minimization for sequence-level knowledge distillation.
\newblock In {\em ACL}, 2023.

\bibitem{tinyclip}
Kan Wu, Houwen Peng, Zhenghong Zhou, Bin Xiao, Mengchen Liu, Lu Yuan, Hong
  Xuan, Michael Valenzuela, Xi~Stephen Chen, Xinggang Wang, Hongyang Chao, and
  Han Hu.
\newblock Tinyclip: Clip distillation via affinity mimicking and weight
  inheritance.
\newblock In {\em ICCV}, 2023.

\bibitem{xie2018improving}
Jiafeng Xie, Bing Shuai, Jian-Fang Hu, Jingyang Lin, and Wei-Shi Zheng.
\newblock Improving fast segmentation with teacher-student learning.
\newblock {\em BMVC}, 2018.

\bibitem{xie2020self}
Qizhe Xie, Minh-Thang Luong, Eduard Hovy, and Quoc~V Le.
\newblock Self-training with noisy student improves imagenet classification.
\newblock In {\em CVPR}, 2020.

\bibitem{yang2023clipkd}
Chuanguang Yang, Zhulin An, Libo Huang, Junyu Bi, Xinqiang Yu, Han Yang, and
  Yongjun Xu.
\newblock Clip-kd: An empirical study of distilling clip models, 2023.

\bibitem{sdft}
Zhaorui Yang, Qian Liu, Tianyu Pang, Han Wang, Haozhe Feng, Minfeng Zhu, and
  Wei Chen.
\newblock Self-distillation bridges distribution gap in language model
  fine-tuning.
\newblock {\em ACL}, 2024.

\bibitem{ye2023mplugowl}
Qinghao Ye, Haiyang Xu, Guohai Xu, Jiabo Ye, Ming Yan, Yiyang Zhou, Junyang
  Wang, Anwen Hu, Pengcheng Shi, Yaya Shi, Chenliang Li, Yuanhong Xu, Hehong
  Chen, Junfeng Tian, Qi Qian, Ji Zhang, Fei Huang, and Jingren Zhou.
\newblock mplug-owl: Modularization empowers large language models with
  multimodality, 2023.

\bibitem{yuan2020revisiting}
Li Yuan, Francis~EH Tay, Guilin Li, Tao Wang, and Jiashi Feng.
\newblock Revisiting knowledge distillation via label smoothing regularization.
\newblock In {\em CVPR}, 2020.

\bibitem{Yuan2023TinyGPTVEM}
Zhengqing Yuan, Zhaoxu Li, and Lichao Sun.
\newblock Tinygpt-v: Efficient multimodal large language model via small
  backbones.
\newblock {\em ArXiv}, 2023.

\bibitem{zhang2020improve}
Linfeng Zhang and Kaisheng Ma.
\newblock Improve object detection with feature-based knowledge distillation:
  Towards accurate and efficient detectors.
\newblock In {\em ICLR}, 2020.

\bibitem{zhang2020distilling}
Yuan Zhang, Xiaoran Xu, Hanning Zhou, and Yan Zhang.
\newblock Distilling structured knowledge into embeddings for explainable and
  accurate recommendation.
\newblock In {\em WSDM}, 2020.

\bibitem{zhu2023minigpt}
Deyao Zhu, Jun Chen, Xiaoqian Shen, Xiang Li, and Mohamed Elhoseiny.
\newblock Minigpt-4: Enhancing vision-language understanding with advanced
  large language models.
\newblock {\em arXiv preprint arXiv:2304.10592}, 2023.

\bibitem{zhu2024llavaphi}
Yichen Zhu, Minjie Zhu, Ning Liu, Zhicai Ou, Xiaofeng Mou, and Jian Tang.
\newblock Llava-phi: Efficient multi-modal assistant with small language model,
  2024.

\bibitem{Zuo2022MoEBERTFB}
Simiao Zuo, Qingru Zhang, Chen Liang, Pengcheng He, Tuo Zhao, and Weizhu Chen.
\newblock Moebert: from bert to mixture-of-experts via importance-guided
  adaptation.
\newblock In {\em NAACL}, 2022.

\end{thebibliography}
}

\clearpage
\appendix
\section{Appendix}
\label{sec:appendix}





\subsection{The instruction-tuning data generated by the teacher model.}
\begin{figure}[h]
    \centering
    \includegraphics[bb=0 0 1097 1281, width=\textwidth]{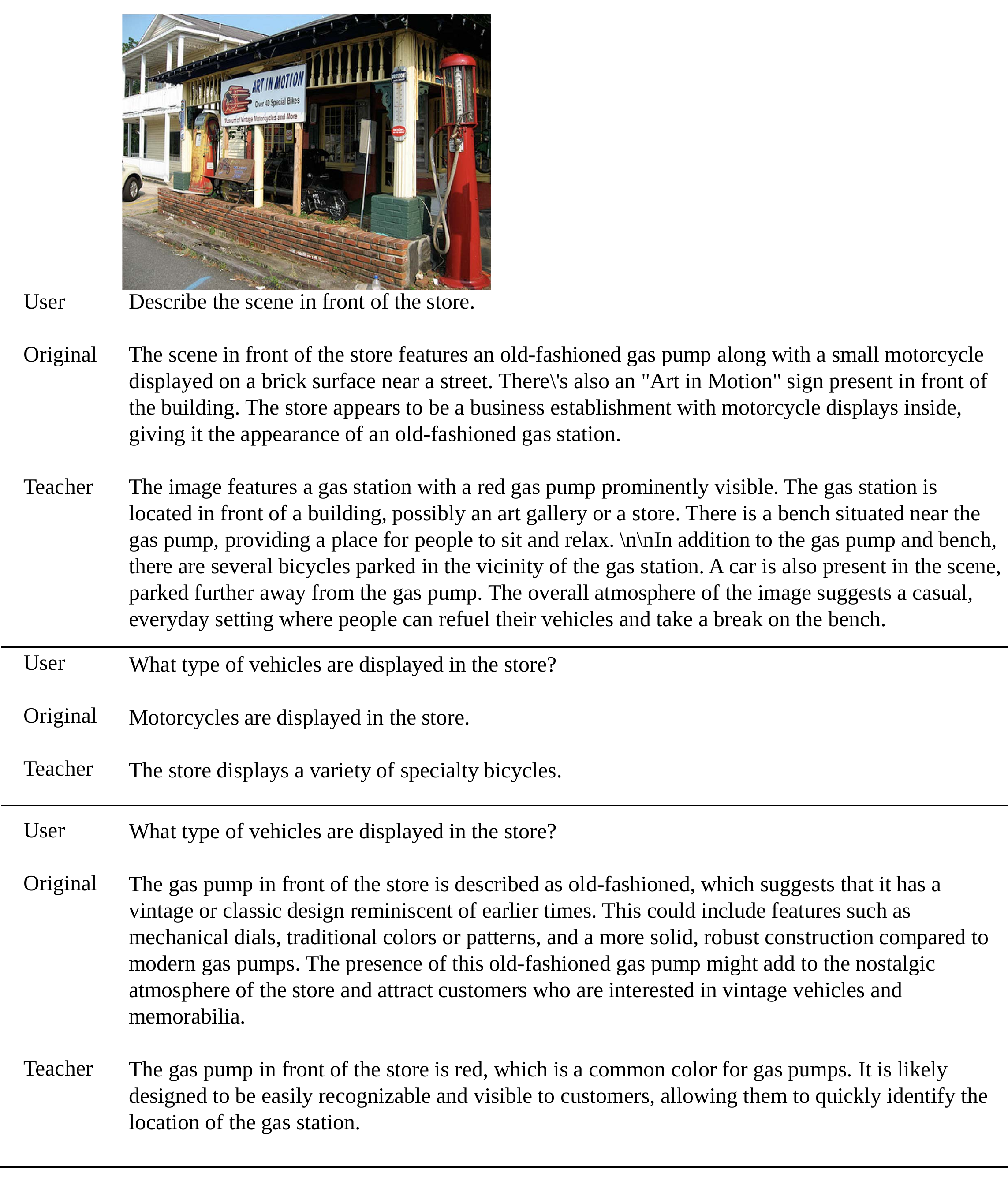}
    \caption{Comparison of original instruction-tuning data and data generated by the teacher model.}
    \label{fig:tea_data}
\end{figure}

The 150K instruction-tuning data for LLaVA~\cite{liu2023llava} was generated by GPT-3.5, which is solely a language model. These datasets contain inaccuracies or hallucinations, potentially impacting the quality and reliability of the resulting models. The data distribution of these datasets may not be compatible with the teacher and student models. So, we try to regenerate the answer to each question using the teacher. So, we try to distill the knowledge from the teacher model to the student model through the data perspective. In Fig.~\ref{fig:tea_data}, we show a conversation example generated by the teacher model. 

\subsubsection{Explore the more fine-tuning dataset.}
To verify the scalability of our distillation strategies, we conduct experiments using a combination of 2M instruction-tuning datasets. This mix includes various types of data, such as VQA, region-level VQA, visual conversation, and caption data. 
This mix contains different types of data, including VQA, region-level VQA, visual conversation, and caption data. 
All these diverse data types are combined and sampled equally. The data is structured in a dialogue format similar to LLaVA!\cite{liu2023llava}.

\begin{table}[h]
    \centering
    \setlength{\tabcolsep}{4mm} {
    \begin{tabular}{c|cc}
    \toprule
    Datasets & Type & Samples \\
    \toprule
    \rowcolor{whitesmoke} \multicolumn{3}{l}{\textit{\textbf{Pretraining}}}   \\
    BLIP-LCS~\cite{li2022blip} & Caption & 558K \\
    \midrule
    \rowcolor{whitesmoke} \multicolumn{3}{l}{\textit{\textbf{Fine-tuning}}}   \\
    Visual Dialog ~\cite{Das_2017} & Conversation & 123K \\
    Text-VQA~\cite{Singh_2019} & VQA & 35K \\
    VSR~\cite{Liu_2023} & VQA & 13K \\
    VIGC~\cite{Wang2023VIGCVI} & VQA & 37K \\
    IconQA~\cite{Lu2021IconQAAN} & VQA & 107K \\
    SQA~\cite{Lu2022LearnTE} & VQA & 13K \\
    COCO~\cite{Chen2015MicrosoftCC} & Caption & 592K \\
    SBU~\cite{ordonez2011im2text} & Caption & 844K \\
    ShareGPT4v~\cite{chen2023sharegpt4v} & mixed & 665K \\
    \toprule
    \end{tabular}}
    \caption{The 2M mixed instruction dataset for fine-tuning includes a "Type" column specifying the task of the dataset. "Mixed" denotes datasets with samples from different tasks. The "Samples" column indicates the number of image-text pairs in each dataset.}
    \label{tab:mix_data}
\end{table}

\section{Hyperparameters}
We employ MobileLLaMA~\cite{chu2023mobilevlm} as our base LLM, utilizing the identical set of hyperparameters as LLaVA~\cite{liu2023llava}. It is important to note that we did not employ any distillation strategy and solely trained the projection layer during pretraining. 

\begin{table}[h]
    \centering
    \setlength{\tabcolsep}{4mm} {
    \begin{tabular}{c|cc}
    \toprule
    Hyperparameter & Pretraining & Fine-tuning \\
    \toprule
    batch size & 256 & 128 \\
    lr & 1e-3 & 2e-5   \\
    lr schedule & \multicolumn{2}{c}{cosine decay } \\
    lr warmup ratio &  \multicolumn{2}{c}{0.03} \\
    epoch &  \multicolumn{2}{c}{1} \\ 
    optimizer & \multicolumn{2}{c}{AdamW} \\
    DeepSpeed stage & 2 & 3 \\
    \toprule
    \end{tabular}
    }
    \caption{Hyperparameters of our method. }
    \label{tab:hyperparameters}
\end{table}

\noindent
\section{Limitations.} 
\label{sec:limit}
Although we have tried different strategies, our results are higher than the baseline. However, they still fall short of our expectations and significantly lag behind the teacher model's performance in most evaluation benchmarks, indicating the need for further research and refinement to bridge this performance gap.
\clearpage

\end{document}